\documentclass[runningheads]{llncs}
\usepackage[T1]{fontenc}
\usepackage{graphicx,verbatim}
\usepackage{xcolor}
\usepackage{hyperref, xcolor}

\usepackage{amssymb, amsmath, amsfonts}
\usepackage{svg}
\usepackage{hyperref}

\begin{document}
\title{Fully Automated Segmentation of Fiber Bundles in Anatomic Tracing Data}
%
% \begin{comment}  %% Removed for anonymized MICCAI 2025 submission
\author{Kyriaki-Margarita Bintsi \inst{1} \and
Ya\"{e}l Balbastre \inst{2} \and
Jingjing Wu \inst{1} \and
Julia F.~Lehman \inst{3} \and
Suzanne N.~Haber \inst{3,4} \and
Anastasia Yendiki \inst{1} 
}
\authorrunning{K. M. Bintsi et al.}
\institute{Athinoula A. Martinos Center for Biomedical Imaging, Massachusetts General Hospital and Harvard Medical School, Charlestown, MA, United States 
\email{kbintsi@mgh.harvard.edu}\\
\and
Department of Experimental Psychology, University College London, London, United Kingdom
\and
Department of Pharmacology and Physiology, University of Rochester School of Medicine, Rochester, NY, United States
\and
McLean Hospital, Belmont, MA, United States}

% \end{comment}

% \author{Anonymized Authors}  %% Added for anonymized MICCAI 2025 submission
% \authorrunning{Anonymized Author et al.}
% \institute{Anonymized Affiliations \\
%     \email{email@anonymized.com}}

\maketitle              % typeset the header of the contribution

\begin{abstract}
% The abstract should briefly summarize the contents of the paper in 150--250 words.  If you are to include a link to your Repository, please make sure it is anonymized for the double-blind review phase.
Anatomic tracer studies are critical for validating and improving diffusion MRI (dMRI) tractography. However, large-scale analysis of data from such studies is hampered by the labor-intensive process of annotating fiber bundles manually on histological slides. Existing automated methods often miss sparse bundles or require complex post-processing across consecutive sections, limiting their flexibility and generalizability. We present a streamlined, fully automated framework for fiber bundle segmentation in macaque tracer data, based on a U-Net architecture with large patch sizes, foreground aware sampling, and semi-supervised pre-training. Our approach eliminates common errors such as mislabeling terminals as bundles, improves detection of sparse bundles by over 20\% and reduces the False Discovery Rate (FDR) by 40\% compared to the state-of-the-art, all while enabling analysis of standalone slices. This new framework will facilitate the automated analysis of anatomic tracing data at a large scale, generating more ground-truth data that can be used to validate and optimize dMRI tractography methods.
\keywords{Anatomic tracing \and fiber bundle segmentation \and semi-supervised learning.}
% Authors must provide keywords and are not allowed to remove this Keyword section.

\end{abstract}
\section{Introduction}
Diffusion Magnetic Resonance Imaging (dMRI) tractography has emerged as a crucial non-invasive technique for in vivo reconstruction of white matter pathways, playing a vital role in understanding psychiatric and neurological disorders \cite{catani2006diffusion,yang2021diffusion}. However, its accuracy is limited by the fact that dMRI relies on indirect measurements of axonal orientations based on water diffusion, and that its millimeter scale resolution is much coarser than the size of individual axons \cite{thomas2014anatomical,grisot2021diffusion}.

In contrast, anatomic tracer studies in non-human primates (NHPs) allow direct visualization of axonal trajectories at the microscale. These studies provide crucial ground-truth data for validating and understanding white matter pathways reconstructed through non-invasive neuroimaging in humans. Comparison between tracer injections and dMRI in the same NHP samples has yielded valuable information on the anatomic accuracy of different dMRI acquisition and analysis methods \cite{safadi2018functional,yendiki2022post}.

Microscopy data from brains that have received anatomic tracer injections allow detailed visualization of axonal projections from the injection sites, as they travel through the white matter to reach their terminals. This high-resolution mapping reveals intricate brain connectivity patterns beyond what can be resolved with dMRI tractography, including how the fibers originate, branch into sub-bundles, and take tortuous routes all the way to their terminations \cite{schilling2019limits,safadi2018functional,jbabdi2013human}. However, a significant bottleneck in leveraging such data lies in the manual annotation of axon bundles in histological slides, which is extremely time- and labor-intensive. This has restricted the availability of annotated data and, consequently, limited large-scale validation studies of dMRI.

Previous work in automating this annotation has been limited \cite{sundaresan2022constrained,sundaresan2025self}. The vast majority of existing axon segmentation methods focus on dense ultra high-resolution images such as those acquired with electron microscopy \cite{wei2021axonem,naito2017identification}. These methods are not applicable to images of whole NHP brain sections acquired at micrometer resolution, where the segmentation task is to separate axon bundles that contain tracers from other sources of signal (injection site, terminals, background signal). In this context, work has been mostly limited to images acquired with light-sheet fluorescence microscopy in mice \cite{winnubst2019reconstruction,friedmann2020mapping}, where imaging is inherently three-dimensional. Segmentation of tracer data has also been performed in marmoset monkeys, using serial two-photon imaging \cite{yan2022mapping} and serial histology \cite{woodward2020nanozoomer}. In both cases, all positively stained pixels were segmented, without differentiating axons from their terminals. Macaques provide a closer homologue to the human brain than marmosets, making them more suitable for translational research \cite{chatterjee2009estimating}. A deep-learning method for segmenting axon bundles in macaque tracer data was proposed in \cite{sundaresan2025self}. Although it is a significant first step towards automating this task, this method relies significantly on post-processing to reduce false positives by comparison of predicted bundles across consecutive sections, hence limiting its applicability to single sections. Additionally, the limited training data used in that work affect the generalizability of the model, as seen, e.g., from its limited accuracy in detecting sparse fiber bundles. Thus, it is suitable for computer-aided, rather than fully automated, segmentation.

Here we propose a semi-supervised framework for fiber bundle segmentation in macaque tracer data. Our approach is based on U-Net architectures enhanced with pre-training strategies, and it is designed to operate effectively with limited manual annotations. We evaluate our method in multiple macaque brains with tracer injections in varying sites, demonstrating its ability to generalize across brains and fiber configurations. Our approach outperforms the state-of-the-art, including a 22\% improvement for sparse fiber bundles.  The code is available on GitHub at: \url{https://github.com/lincbrain/fiber-bundle-segmentation}.

\section{Methods}
\begin{figure}
\includegraphics[width=\textwidth]{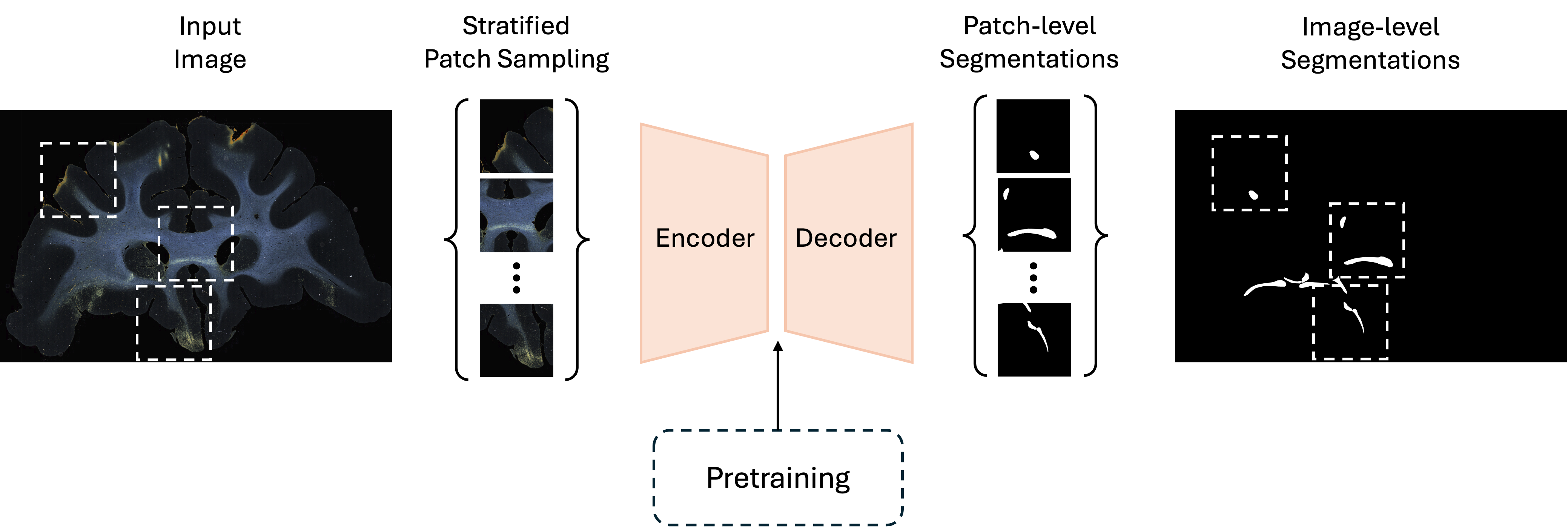}
\caption{Proposed pipeline. Patches are extracted from the images using a stratified patch sampling technique that ensures foreground is available to 50\% of the patches.  These patches are then used to train a U-Net. Patch-level segmentations are combined back to image-level. Pre-training of the U-Net for a reconstruction task is used to leverage the availability of unlabeled images.} \label{fig1}
\end{figure}
An overview of the pipeline can be found in Figure \ref{fig1}.

\subsection{Patch sampling and pre-processing}

Due to the large size of the histological images and the sparsity of annotated fiber bundles, we adopt a patch-based training strategy. Each whole-slide image is downsampled (see Section~\ref{dataset}) and divided into smaller patches for model training and inference.

We apply Z-score normalization to stabilize training. During training, we extract 20 random patches of size $1024 \times 1024$ pixels from each section. We deploy a stratified patch sampling approach, which ensures that 50\% of the sampled patches contain foreground (i.e., fiber bundles), thereby balancing background and foreground regions.

To account for limited labeled data, we use a combination of data augmentation strategies — horizontal and vertical flips and elastic transformations.

\subsection{Model architecture and training}

We employ a 2D U-Net architecture~\cite{ronneberger2015u} for fiber bundle segmentation. Our configuration for network training involves a combination of Binary Cross Entropy (BCE) and Dice loss. 

Additionally, we consider an alternative setup using a combination of Focal Loss~\cite{lin2017focal} and Dice Loss, designed to address class imbalance and enhance overlap with the ground truth.

The total loss function is defined as:
\begin{equation}
\mathcal{L}_{\text{total}} = \mathcal{L}_{\text{BCE}} + \mathcal{L}_{\text{Dice}} \quad
\text{or }  \quad  \mathcal{L}_{\text{total}} =\mathcal{L}_{\text{Focal}} + \mathcal{L}_{\text{Dice}},
\end{equation}
depending on the configuration. 
The BCE loss is given by:
\begin{equation}
\mathcal{L}_{\text{BCE}} = -\frac{1}{N} \sum_{i=1}^{N} \left[ g_i \log(p_i) + (1 - g_i) \log(1 - p_i) \right],
\end{equation}
where $p_i$ and $g_i$ denote the predicted and ground truth probabilities for pixel $i$, respectively. 
The Focal Loss is defined as:
\begin{equation}
\mathcal{L}_{\text{Focal}} = -\frac{1}{N} \sum_{i=1}^{N} \alpha_t (1 - p_t)^\gamma \log(p_t),
\end{equation}
where $p_t$ is the predicted probability for the true class, and $\alpha_t$, $\gamma$ are hyperparameters used to balance class weights and focus training on hard examples. 
The Dice Loss is:
\begin{equation}
\mathcal{L}_{\text{Dice}} = 1 - \frac{2 \sum_i p_i g_i}{\sum_i p_i + \sum_i g_i + \epsilon},
\end{equation}
with $\epsilon$ added to prevent division by zero.

\subsection{Inference and postprocessing}

At inference time, the model processes each image using a sliding window approach with a stride of 25\% of the patch size. The resulting probability maps from overlapping patches are averaged to form a final prediction map. Post-processing steps include Gaussian smoothing, mirroring, and removal of very small connected components that likely represent noise.

\subsection{Self-supervised pre-training}

While manually labeled sections are scarce, unlabeled sections are plentiful. To leverage the latter, we investigate whether self-supervised pre-training improves downstream segmentation performance. Specifically, we pre-train the U-Net on an image reconstruction task. The network is trained to reconstruct an input histological patch from a compressed latent representation, encouraging the encoder to learn meaningful structural features of tracer data.

We use Mean Squared Error (MSE) as the reconstruction loss:
\begin{equation}
\mathcal{L}_{\text{MSE}} = \frac{1}{N} \sum_{i=1}^{N} (x_i - \hat{x}_i)^2,
\end{equation}
where $x_i$ and $\hat{x}_i$ are the pixel values of the input and reconstructed image, respectively, and $N$ is the total number of pixels in a patch.

The learned weights are then used to initialize the segmentation network, which is subsequently fine-tuned using labeled data.

\section{Experiments}

\subsection{Datasets}
\label{dataset}

We used high-resolution coronal histological sections of macaque brains with injections of bidirectional tracers at different cortical sites. In total, data from $N=20$ macaques were included, with injection sites at distinct anatomical areas, resulting in a diverse set of axonal trajectories and fiber configurations.

The histological sections had a thickness of 50~{\textmu}m, and were digitized with an in-plane resolution of 0.4~{\textmu}m. To reduce redundancy and processing load while maintaining anatomical continuity, every eighth section was stained to visualize a specific tracer, resulting in an effective inter-slice gap of 400~{\textmu}m. This sampling strategy follows previous tracer studies in non-human primates \cite{lehman2011rules,haynes2013organization}.

Due to the labor-intensive nature of manual annotations, expert-labeled ground truth is available for a limited subset of the data: five complete macaque brains (M1-M5). 
%and a small number of additional slices from a sixth (M6). 
All annotations were performed by an expert neuroanatomist and include segmentation of three classes of fiber bundles: dense, moderate, and sparse. A fiber bundle is defined here as a group of axons that travel in close proximity to each other, from the injection site toward one or more terminal fields.

Because the original 2D histological images were extremely large, we downsampled all images by a factor of 4 prior to training and evaluation, yielding a resolution suitable for processing while preserving key anatomical features.

In total, we collected 898 histological sections (labeled and unlabeled) from macaques M1–M20, of which 165 sections had manually annotated fiber bundles. For model training and validation, we used the complete data set of macaque M1 and the anterior and posterior halves of sections from macaques M2 and M3, respectively. For model testing, we used the left-out halves of M2 and M3 sections, and the complete datasets of M4 and M5, allowing us to evaluate the generalization of the model to unseen brains and fiber trajectories.

\subsection{Evaluation metrics}
\label{metrics}
We follow the evaluation protocol from \cite{sundaresan2025self} to allow for a direct comparison of
performance.

\noindent \textbf{True positive rate; TPR (Sensitivity).} We quantify sensitivity by the TPR, which measures the proportion of fiber bundles in the ground truth that are correctly detected by the model. We report sensitivity separately for each fiber bundle class: dense, moderate, and sparse. Note that our model does not discriminate between these classes during training or inference, but evaluating performance by class is informative, as will be shown below. The TPR for each class is computed as:
\begin{equation}
\text{TPR}_{\text{class}} = \frac{\text{TP}_{\text{class}}}{\text{TP}_{\text{class}} + \text{FN}_{\text{class}}}, 
\end{equation}
where $\text{TP}_{\text{class}}$ denotes the number of ground truth bundles of a given class that are successfully matched by a predicted bundle, and $\text{FN}_{\text{class}}$ are ground truth bundles of that class with no corresponding prediction. Bundle-wise matching is determined via connected component analysis.

\noindent \textbf{Average number of true and false positives per section ($\text{TP}_\text{avg}$, $\text{FP}_\text{avg}$).} These metrics represent the average number of predicted fiber bundles per test section that correctly (TP) or incorrectly (FP) overlap with ground truth annotations. A predicted bundle is counted as a true positive if it overlaps with any ground truth bundle, and as a false positive otherwise. The average is computed by dividing the total TP and FP counts by the number of test sections.

\noindent \textbf{False discovery rate (FDR).} The FDR quantifies the proportion of incorrect positive predictions among all positive predictions made by the model, with a lower FDR indicating fewer false positives. FDR is given by:
\begin{equation}
\text{FDR} = \frac{\text{FP}}{\text{TP} + \text{FP}},
\end{equation}
where TP and FP are computed as described above using component-wise matching between predictions and ground truth.

\subsection{Implementation details}
We employ a flexible 2D U-Net architecture~\cite{ronneberger2015u}. The network consists of an encoder and decoder with skip connections. The number of feature maps doubles after each downsampling block in the encoder, starting from a base of 32 and capped at 512. Specifically, we have nine resolution levels in total. The decoder performs upconvolution followed by concatenation with the corresponding encoder output and a double convolution block.
ReLU is used as the activation function in all convolutional blocks. 
% In the focal loss, we set $\alpha_t = 0.25$ and $\gamma = 2.0$, following the original formulation in~\cite{lin2017focal}.
The network is trained with the Adam optimizer  \cite{kingma2014adam} and a base learning rate of \(10^{-4}\) for 1000 epochs.
To maximize robustness and minimize variance, we performed 5-fold cross-validation. The final prediction for each patch was obtained by averaging the scores from all 5 models. During inference, we applied a grid sampling strategy with overlapping patches (using a stride of 0.25 of the patch size) and then blended overlapping regions to produce the final segmentation. The pipeline was implemented in PyTorch \cite{paszke2019pytorch} and trained on
a Titan RTX GPU. 

\begin{figure}[t]
\includegraphics[width=\textwidth]{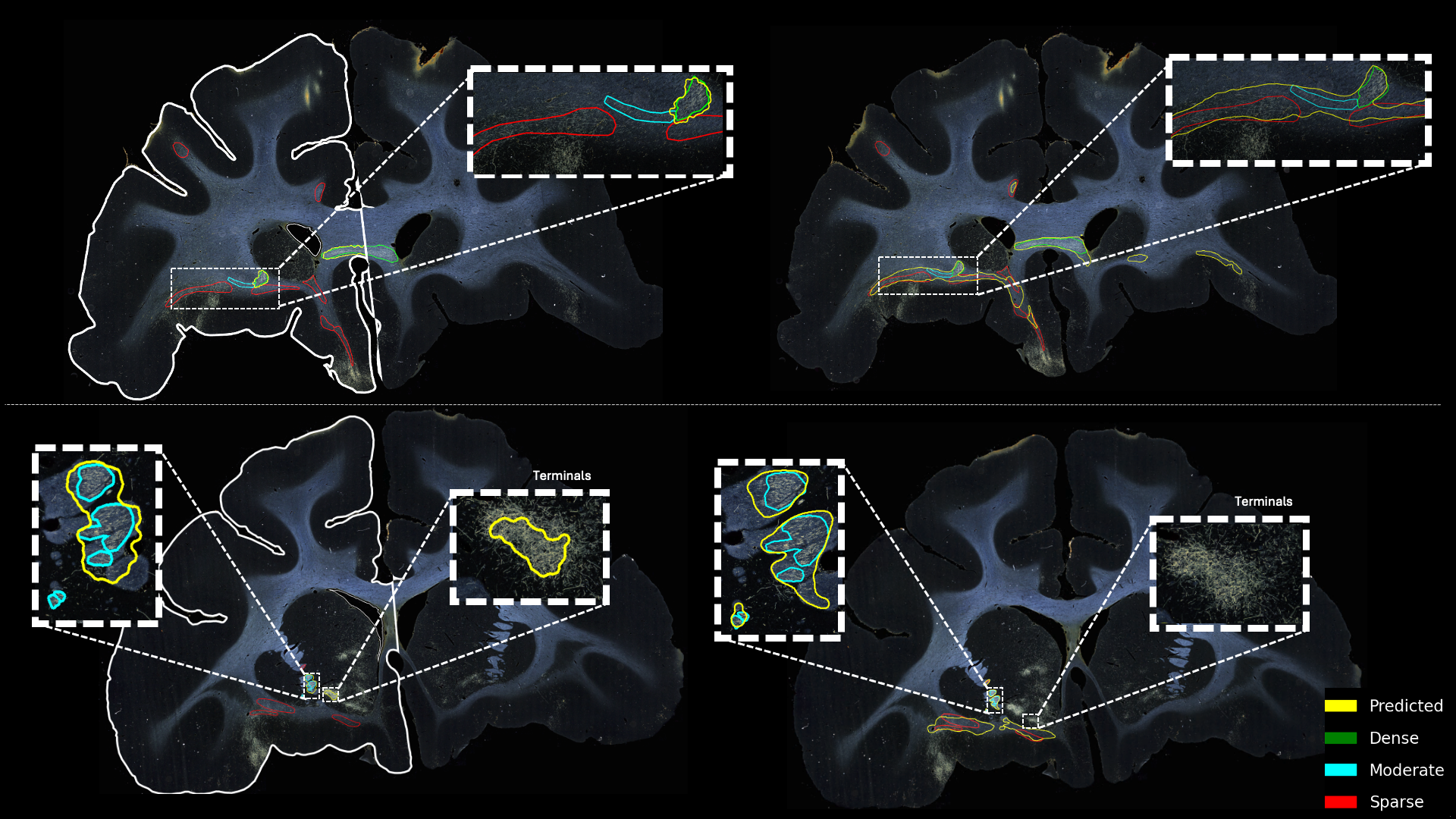}
\caption{Visual comparison between \textbf{the state-of-the-art method \cite{sundaresan2025self} (left)} and \textbf{our approach (right)} on two histological sections.
Predicted segmentations are shown in yellow. Ground truth annotations indicate dense bundles in green, moderate bundles in cyan, and sparse bundles in red.
Our approach is able to detect sparse bundles that the baseline misses (\textbf{top row}), while also correctly separating terminals from fiber bundle areas (\textbf{bottom row}).
 Note that we have ground truth annotations only inside the white outline (ipsilateral hemisphere), hence predictions made outside this outline are not included in the evaluation metrics.
} \label{fig2}
\end{figure}

\section{Results}

To provide a baseline for comparison, we first evaluated the state-of-the-art~\cite{sundaresan2025self} on our test set. We then trained four variations of our model: 1) using the exact same training dataset as in \cite{sundaresan2025self}, which we refer to as $D_\mathrm{small}$; 2) using the larger training set described in Section \ref{dataset}, which we refer to as $D_\mathrm{large}$, but including only dense and moderate bundles as was done in \cite{sundaresan2025self}; 3) using the larger training set and including dense, moderate, and sparse bundles, which we refer to as $D_\mathrm{large+sparse}$; and 4) repeating (3) with the use of pre-training.
The test set consists of the annotated sections from brains M2 and M3 that were not used during training, as well as all annotated sections from two additional brains, M4 and M5. The results are summarized in Table \ref{tab:overall}. 

\begin{table}[t]
\caption{Performance of the proposed approach, with the state-of-the-art \cite{sundaresan2025self} for fiber bundle segmentation from anatomic tracing data acting as a baseline for comparison. Explanations of the datasets and the metrics can be found in Section \ref{dataset} and Section \ref{metrics} respectively.}
\centering
\begin{tabular}{l|c|c|c|c|c}
\hline
\textbf{Model} & \textbf{Training}  & \textbf{TPR} $\uparrow$  & \textbf{TP$_\text{avg}$} & \textbf{FP$_\text{avg}$} &\textbf{FDR} \\ 
 & \textbf{dataset} &  \textbf{dense/moderate/sparse} &  $\uparrow$  & $\downarrow$ & $\downarrow$ \\
\hline
Baseline \cite{sundaresan2025self} & $D_\mathrm{small}$ & 0.93/0.80/0.49 & 1.00 & 3.07 & 0.63 \\ 
Ours (1) & $D_\mathrm{small}$  & 0.92/0.87/0.60 & 1.82 & 1.08
 & 0.24 \\ 
Ours (2) & $D_\mathrm{large}$ & 0.97/0.85/0.66  & 2.54 & 0.66 & 0.16 \\ 
Ours (3) & $D_\mathrm{large+sparse}$ & 0.98/0.94/0.74 & 2.81 & 1.37 & 0.28 \\ 
Ours+pre-train (4) & $D_\mathrm{large+sparse}$ & 0.98/0.94/0.71 & 3.11 & 1.13 & 0.19 \\ 
\hline
\end{tabular}
\label{tab:overall}
\end{table}

Our pipeline performed best when leveraging pre-training and the $D_\mathrm{large+sparse}$ dataset. It demonstrated strong performance across all bundles, yielding TPR$_\text{dense}$ = 0.98, TPR$_\text{moderate}$ = 0.94, and TPR$_\text{sparse}$ = 0.71, while increasing true positives (TP$_\text{avg}$ = 3.11) and reducing low FP$_\text{avg}$ at 1.13 and FDR at 0.19. Importantly, it performed considerably better than the baseline \cite{sundaresan2025self} (TPR$_\text{sparse}$ at 0.49 and FDR at 0.63).

Qualitative comparisons between our best configuration and the baseline \cite{sundaresan2025self} can be found in Figure \ref{fig2}. The baseline method detects dense bundles but misses several moderate and most sparse ones (Figure \ref{fig2} \textbf{top row}). In contrast, our method correctly identifies bundles across all density levels.
Additionally, while the baseline predicts some moderate bundles, it largely misses sparse ones and erroneously labels terminals as fiber areas (Figure \ref{fig2} \textbf{bottom row}). Our model resolves both issues, producing more complete and accurate segmentations. Note here that manual annotations were available only within the brain outlines shown in the figures, which typically included only the ipsilateral hemisphere of the injection site. As we did not have ground truth labels outside that outline, we excluded any predictions outside the outline when computing evaluation metrics, for both our models and the baseline model (seen on the left of Figure \ref{fig2}).  It is important to note, however, that predictions outside the outline are not necessarily false positives, as we do expect the injection sites to project to the contralateral hemisphere.

\subsection{Ablation studies}

\textbf{Expansion of the training set.}
This comparison highlights how additional manually annotated training data influence performance. Comparing the U-Net trained on $D_\mathrm{small}$ (Table \ref{tab:overall}:(1)), that is, the same training data distribution as the baseline~\cite{sundaresan2025self}, we notice that our model improves performance, with an almost 3x reduction in FP$_\text{avg}$ from 3.07 to only 1.08.
Leveraging more data with $D_\mathrm{large}$ (Table \ref{tab:overall}:(2)), even without including sparse bundles, already leads to improvement — yielding TPR$_\text{dense}$ of 0.97 (vs.~0.92) and TPR$_\text{sparse}$ of 0.66 (vs.~0.60) — while increasing TP$_\text{avg}$ from 1.82 to 2.54 and reducing FDR from 0.24 to 0.16. 

\noindent \textbf{Inclusion of sparse bundles during training.}
Here we observe the effects of adding sparse bundles to the training set (Table \ref{tab:overall}:(3)). Our network (without pre-training) performs better at detecting sparse bundles — TPR$_\text{sparse}$ jumped from 0.66 to 0.74 — suggesting that including them in the training set is crucial for generalizability.

\noindent \textbf{Pre-training.}
Our results indicate that pre-training (Table \ref{tab:overall}:(4)) leads to improved model precision, as seen in the lower FDR (0.19 from 0.3), and increases the average number of true positives. However, this comes at the cost of a a slightly reduced sensitivity for sparse bundles to TPR$_\text{sparse}$ of 0.71 (vs.~0.74). This suggests that reconstruction pretraining may instill a conservative bias in the model's feature representation, making it less responsive to weak foreground signals even after fine-tuning.

\noindent \textbf{Loss functions.}
To account for class imbalance, we experimented with Focal+Dice loss instead of the traditional BCE+Dice loss under both pretrained and non-pretrained settings. Without pretraining, focal loss improved the true positive average (TP$\text{avg}$: 2.81 vs. 2.65), particularly enhancing recall in sparse fiber bundles. However, this also led to a slightly elevated false discovery rate (FDR: 0.28 vs. 0.26). When pretraining was applied, focal loss continued to slightly improve TP$_\text{avg}$ while achieving a lower FDR (0.19 vs. 0.21), indicating that pretraining mitigates the instability often associated with focal loss and leads to better precision-recall tradeoffs overall. We attribute the improved performance of focal loss to its ability to down-weight well-classified (easy) negatives, allowing the model to focus on harder, underrepresented positive examples — a property particularly beneficial in the context of sparse or imbalanced fiber bundle distributions.

\section{Discussion}
Fiber bundle segmentation from anatomic tracing data is challenging due to the scarcity of manually labeled ground truth. Our work achieves significant performance improvements, taking an important step toward fully automated segmentation and enabling more effective use of tracer data to validate and optimize dMRI tractography.

Our main contribution is to demonstrate that a simplified architecture, using large patch sizes, foreground-aware sampling, and inclusion of sparse bundles during training can outperform a previous, more complex model that relied on temporal ensembling and contextual information from consecutive sections \cite{sundaresan2025self}. Our method does not require information from neighboring sections, and can thus work even on standalone sections from new datasets. While individual metric differences may appear subtle, the combined improvements—such as a lower FDR alongside a higher TPR for sparse bundles—reflect a clearer and more consistent enhancement in segmentation performance. By reducing architectural complexity while improving accuracy across all evaluation metrics, we establish a new, strong baseline for future work.

A crucial advancement is our method’s ability to accurately detect fiber bundles across the full spectrum of density, including sparse bundles that were previously missed. Additionally, our approach corrects prior mislabeling of terminals as bundles (see Figure \ref{fig2}), resulting in more precise connectivity mapping. This performance improvement is essential for minimizing the need for manual intervention and moving closer to fully automated segmentation. This will facilitate the processing of data from anatomic tracer studies, towards the ultimate goal of building more complete models of brain circuitry and  expanding the availability of ground-truth data to validate dMRI findings.

Although pre-training improves precision, it appears to slightly reduce model sensitivity, particularly in detecting sparse bundles. We hypothesize that this is due to a conservative prediction bias introduced during pre-training, which persists even with full fine-tuning. In future work, we plan to address this limitation by incorporating synthetic training data to increase the diversity and frequency of sparse bundle examples that we present to the network. 
Additionally, we will explore transformer-based models and masked autoencoders as an alternative pre-training strategy that may encourage more informative and task-relevant feature representations compared to standard image reconstruction. To provide a more comprehensive evaluation of model performance, we aim to include additional object- and pixel-level metrics (e.g., Dice coefficient, IoU, detection rate) in future work. These will help characterize model behavior more fully and facilitate better comparisons across methods. We also recognize the importance of reporting variance across runs to assess robustness, and we plan to incorporate such estimates in subsequent studies.
Finally, we plan to extend our model to 3D, so that we can apply it to volumetric imaging data from light-sheet fluorescence microscopy.

\section{Conclusions}
In this work, we present a framework for automated segmentation of fiber bundles in macaque tracer data that surpasses the state-of-the-art in both accuracy and robustness. Specifically, our framework improves segmentation of sparse bundles and allows single-section analysis. By leveraging simple yet effective architectural and sampling strategies, our approach resolves key limitations of prior work—substantially reducing false positives without relying on complex post-processing or contextual information from neighboring sections. This new baseline will enable broader, more scalable extraction of ground-truth data to validate non-invasive imaging of brain connectivity, and will, more generally, support the transition to fully digitized anatomy studies.

% \begin{comment}  %% removed for anonymized MICCAI 2025 submission.
    
    % The following acknowledgement and disclaimer sections should be removed for the double-blind review process.  
    % If and when your paper is accepted, reinsert the acknowledgement and the disclaimer clause in your final camera-ready version.

\begin{credits}
\subsubsection{\ackname} 
This work was supported by the center for Large-scale Imaging of Neural Circuits (LINC), an NIH BRAIN Initiative Connectivity across Scales (CONNECTS) comprehensive center (UM1-NS132358). Additional support was provided by the National Institute for Mental Health (R01-MH045573, P50-MH106435) and the National Institute for Neurological Disorders and Stroke (R01-NS119911, R01-NS127353).

\subsubsection{\discintname}
The authors have no competing interests to declare that are
relevant to the content of this article. 
\end{credits}

% \end{comment}

% \newpage
\bibliographystyle{splncs04}
\bibliography{bibliography}

\begin{thebibliography}{10}
\providecommand{\url}[1]{\texttt{#1}}
\providecommand{\urlprefix}{URL }
\providecommand{\doi}[1]{https://doi.org/#1}

\bibitem{catani2006diffusion}
Catani, M.: Diffusion tensor magnetic resonance imaging tractography in cognitive disorders. Current opinion in neurology  \textbf{19}(6),  599--606 (2006)

\bibitem{chatterjee2009estimating}
Chatterjee, H.J., Ho, S.Y., Barnes, I., Groves, C.: Estimating the phylogeny and divergence times of primates using a supermatrix approach. BMC evolutionary biology  \textbf{9},  1--19 (2009)

\bibitem{friedmann2020mapping}
Friedmann, D., Pun, A., Adams, E.L., Lui, J.H., Kebschull, J.M., Grutzner, S.M., Castagnola, C., Tessier-Lavigne, M., Luo, L.: Mapping mesoscale axonal projections in the mouse brain using a 3d convolutional network. Proceedings of the National Academy of Sciences  \textbf{117}(20),  11068--11075 (2020)

\bibitem{grisot2021diffusion}
Grisot, G., Haber, S.N., Yendiki, A.: Diffusion mri and anatomic tracing in the same brain reveal common failure modes of tractography. Neuroimage  \textbf{239},  118300 (2021)

\bibitem{haynes2013organization}
Haynes, W.I., Haber, S.N.: The organization of prefrontal-subthalamic inputs in primates provides an anatomical substrate for both functional specificity and integration: implications for basal ganglia models and deep brain stimulation. Journal of Neuroscience  \textbf{33}(11),  4804--4814 (2013)

\bibitem{jbabdi2013human}
Jbabdi, S., Lehman, J.F., Haber, S.N., Behrens, T.E.: Human and monkey ventral prefrontal fibers use the same organizational principles to reach their targets: tracing versus tractography. Journal of Neuroscience  \textbf{33}(7),  3190--3201 (2013)

\bibitem{kingma2014adam}
Kingma, D.P., Ba, J.: Adam: A method for stochastic optimization. arXiv preprint arXiv:1412.6980  (2014)

\bibitem{lehman2011rules}
Lehman, J.F., Greenberg, B.D., McIntyre, C.C., Rasmussen, S.A., Haber, S.N.: Rules ventral prefrontal cortical axons use to reach their targets: implications for diffusion tensor imaging tractography and deep brain stimulation for psychiatric illness. Journal of Neuroscience  \textbf{31}(28),  10392--10402 (2011)

\bibitem{lin2017focal}
Lin, T.Y., Goyal, P., Girshick, R., He, K., Doll{\'a}r, P.: Focal loss for dense object detection. In: Proceedings of the IEEE international conference on computer vision. pp. 2980--2988 (2017)

\bibitem{naito2017identification}
Naito, T., Nagashima, Y., Taira, K., Uchio, N., Tsuji, S., Shimizu, J.: Identification and segmentation of myelinated nerve fibers in a cross-sectional optical microscopic image using a deep learning model. Journal of neuroscience methods  \textbf{291},  141--149 (2017)

\bibitem{paszke2019pytorch}
Paszke, A.: Pytorch: An imperative style, high-performance deep learning library. arXiv preprint arXiv:1912.01703  (2019)

\bibitem{ronneberger2015u}
Ronneberger, O., Fischer, P., Brox, T.: U-net: Convolutional networks for biomedical image segmentation. In: Medical image computing and computer-assisted intervention--MICCAI 2015: 18th international conference, Munich, Germany, October 5-9, 2015, proceedings, part III 18. pp. 234--241. Springer (2015)

\bibitem{safadi2018functional}
Safadi, Z., Grisot, G., Jbabdi, S., Behrens, T.E., Heilbronner, S.R., McLaughlin, N.C., Mandeville, J., Versace, A., Phillips, M.L., Lehman, J.F., et~al.: Functional segmentation of the anterior limb of the internal capsule: linking white matter abnormalities to specific connections. Journal of Neuroscience  \textbf{38}(8),  2106--2117 (2018)

\bibitem{schilling2019limits}
Schilling, K.G., Nath, V., Hansen, C., Parvathaneni, P., Blaber, J., Gao, Y., Neher, P., Aydogan, D.B., Shi, Y., Ocampo-Pineda, M., et~al.: Limits to anatomical accuracy of diffusion tractography using modern approaches. Neuroimage  \textbf{185},  1--11 (2019)

\bibitem{sundaresan2022constrained}
Sundaresan, V., Lehman, J.F., Fitzgibbon, S., Jbabdi, S., Haber, S.N., Yendiki, A.: Constrained self-supervised method with temporal ensembling for fiber bundle detection on anatomic tracing data. In: International Workshop on Medical Optical Imaging and Virtual Microscopy Image Analysis. pp. 115--125. Springer (2022)

\bibitem{sundaresan2025self}
Sundaresan, V., Lehman, J.F., Maffei, C., Haber, S.N., Yendiki, A.: Self-supervised segmentation and characterization of fiber bundles in anatomic tracing data. Imaging Neuroscience  \textbf{3},  imag\_a\_00514 (2025)

\bibitem{thomas2014anatomical}
Thomas, C., Ye, F.Q., Irfanoglu, M.O., Modi, P., Saleem, K.S., Leopold, D.A., Pierpaoli, C.: Anatomical accuracy of brain connections derived from diffusion mri tractography is inherently limited. Proceedings of the National Academy of Sciences  \textbf{111}(46),  16574--16579 (2014)

\bibitem{wei2021axonem}
Wei, D., Lee, K., Li, H., Lu, R., Bae, J.A., Liu, Z., Zhang, L., dos Santos, M., Lin, Z., Uram, T., et~al.: Axonem dataset: 3d axon instance segmentation of brain cortical regions. In: Medical Image Computing and Computer Assisted Intervention--MICCAI 2021: 24th International Conference, Strasbourg, France, September 27--October 1, 2021, Proceedings, Part I 24. pp. 175--185. Springer (2021)

\bibitem{winnubst2019reconstruction}
Winnubst, J., Bas, E., Ferreira, T.A., Wu, Z., Economo, M.N., Edson, P., Arthur, B.J., Bruns, C., Rokicki, K., Schauder, D., et~al.: Reconstruction of 1,000 projection neurons reveals new cell types and organization of long-range connectivity in the mouse brain. Cell  \textbf{179}(1),  268--281 (2019)

\bibitem{woodward2020nanozoomer}
Woodward, A., Gong, R., Abe, H., Nakae, K., Hata, J., Skibbe, H., Yamaguchi, Y., Ishii, S., Okano, H., Yamamori, T., et~al.: The nanozoomer artificial intelligence connectomics pipeline for tracer injection studies of the marmoset brain. Brain Structure and Function  \textbf{225},  1225--1243 (2020)

\bibitem{yan2022mapping}
Yan, M., Yu, W., Lv, Q., Lv, Q., Bo, T., Chen, X., Liu, Y., Zhan, Y., Yan, S., Shen, X., et~al.: Mapping brain-wide excitatory projectome of primate prefrontal cortex at submicron resolution and comparison with diffusion tractography. Elife  \textbf{11},  e72534 (2022)

\bibitem{yang2021diffusion}
Yang, J.Y.M., Yeh, C.H., Poupon, C., Calamante, F.: Diffusion mri tractography for neurosurgery: the basics, current state, technical reliability and challenges. Physics in Medicine \& Biology  \textbf{66}(15),  15TR01 (2021)

\bibitem{yendiki2022post}
Yendiki, A., Aggarwal, M., Axer, M., Howard, A.F., van Walsum, A.M.v.C., Haber, S.N.: Post mortem mapping of connectional anatomy for the validation of diffusion mri. Neuroimage  \textbf{256},  119146 (2022)

\end{thebibliography}

\end{document}